\documentclass[conference]{IEEEtran}
\IEEEoverridecommandlockouts
\usepackage{cite}
\usepackage{amsmath,amssymb,amsfonts}
\usepackage{algorithmic}
\usepackage{graphicx}
\usepackage{multirow}
\usepackage{textcomp}
\usepackage{marvosym}
\usepackage{hyperref}
\usepackage{xcolor}
\def\BibTeX{{\rm B\kern-.05em{\sc i\kern-.025em b}\kern-.08em
    T\kern-.1667em\lower.7ex\hbox{E}\kern-.125emX}}
\begin{document}

\title{DNPM: A Neural Parametric Model for the Synthesis of Facial Geometric Details\\

\thanks{* denotes corresponding authors. This work was partially supported by Natural Science Foundation of Fujian Province of China (No. 2023J05001), Natural Science Foundation of Xiamen, China (No. 3502Z20227012), NSFC (No. 62077039) and the Fundamental Research Funds for the Central Universities, China (No. 20720230106).}
}

\DeclareRobustCommand*{\IEEEauthorrefmark}[1]{%
    \raisebox{0pt}[0pt][0pt]{\textsuperscript{\footnotesize\ensuremath{#1}}}}
\author{\IEEEauthorblockN{Haitao Cao\IEEEauthorrefmark{1},
Baoping Cheng\IEEEauthorrefmark{2},
Qiran Pu\IEEEauthorrefmark{3}, 
Haocheng Zhang\IEEEauthorrefmark{4}, 
Bin Luo\IEEEauthorrefmark{1},
Yixiang Zhuang\IEEEauthorrefmark{1},
\\
Juncong Lin\IEEEauthorrefmark{1},
Liyan Chen\IEEEauthorrefmark{1}*,
Xuan Cheng\IEEEauthorrefmark{1}*
}
\IEEEauthorblockA{\IEEEauthorrefmark{1}School of Informatics,
Xiamen University, China}
\IEEEauthorblockA{\IEEEauthorrefmark{2}Department of Electronic Engineering, Tsinghua University, China}
\IEEEauthorblockA{\IEEEauthorrefmark{3}China Mobile (Hangzhou) Information Technology Co., Ltd., China}
\IEEEauthorblockA{\IEEEauthorrefmark{4}School of Computing and Data Sciences, Xiamen University Malaysia, Malaysia}
\IEEEauthorblockA{chtao@stu.xmu.edu.cn, \{chenliyan, chengxuan\}@xmu.edu.cn}
}

\maketitle

\begin{abstract}
Parametric 3D models have enabled a wide variety of computer vision and graphics tasks, such as modeling human faces, bodies and hands. In 3D face modeling, 3DMM is the most widely used parametric model, but can't generate fine geometric details solely from identity and expression inputs. To tackle this limitation, we propose a neural parametric model named DNPM for the facial geometric details, which utilizes deep neural network to extract latent codes from facial displacement maps encoding details and wrinkles. Built upon DNPM, a novel 3DMM named Detailed3DMM is proposed, which augments traditional 3DMMs by including the synthesis of facial details only from the identity and expression inputs. Moreover, we show that DNPM and Detailed3DMM can facilitate two downstream applications: speech-driven detailed 3D facial animation and 3D face reconstruction from a degraded image. Extensive experiments have shown the usefulness of DNPM and Detailed3DMM, and the progressiveness of two proposed applications. The project page is:\href{https://xmuchtao.github.io/DNPM.github.io/}{https://xmuchtao.github.io/DNPM.github.io/}
\end{abstract}


\section{Introduction}
The digitalization of 3D human faces has been a hot research topic in computer graphics and computer vision due to its wide application in film, video games, mixed reality, etc. The academic and industry communities started very early in building a parametric model for representing 3D faces, intending to control the high-dimensional 3D face model by the low-dimensional semantic parameters, e.g. identity and expression. 3D Morphable Model (3DMM) \cite{blanz1999morphable} may be the most widely used parametric model in face-related applications, due to its effective control of facial shape, expression and texture. 

However, 3DMM can only generate a coarse face shape, e.g. FaceWarehouse \cite{facewarehouse} and FLAME \cite{flame}, with the lack of medium and fine-scale geometric details of individuals. There have been attempts, e.g. FaceScape \cite{yang2020facescape}, DECA \cite{DECA} and DCT \cite{DCT}, which can predict geometric details from RGB images, but the geometric details still can't be generated from the inputted low-dimensional parameters in these methods. The recent method FaceVerse \cite{faceverse} proposes a latent code-controllable geometric details synthesis network, but it relies on geometry and texture UV maps as additional inputs. To conclude, \emph{a 3DMM which can produce both coarse 3D face and facial geometric details only from the low-dimensional semantic parameters input, e.g. identity and expression, has not been proposed yet}.

\begin{figure}
\centering
\includegraphics[width=0.41\textwidth]{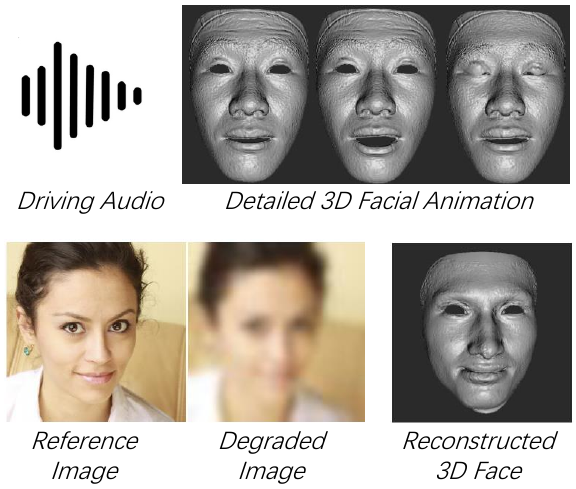}
\caption{The proposed DNPM and Detailed3DMM can enable detailed 3D facial animation from driving audio, and detailed 3D face reconstruction from degraded facial image.}

\label{fig:banner}
\end{figure}

To tackle this limitation, in this paper, we construct a parametric model for facial geometric details, which can further enable semantic-controllable facial details synthesis. Instead of modeling in the 3D space, we conduct parameterization in the 2D space, with the considerations in dataset and methodology. On the one hand, current large-scale 3D face datasets, e.g. FaceScape \cite{yang2020facescape}, usually use the 3D base mesh and 2D displacement map to represent a detailed 3D face model, whereby the 3D wrinkles are well expressed by the 2D image. On the other hand, several deep neural networks proposed in the image generation field can be utilized to compact data and then extract latent code for parametric representation. 
Based on the above two observations, we train a StyleGAN v2 model \cite{karras2020analyzing} over the displacement maps from the FaceScape dataset, thereby representing a $1024\times1024$ displacement map by a $18\times512$ dimension vector. We name the trained StyleGAN v2 model on facial geometric details as \emph{D}etails' \emph{N}eural \emph{P}arametric \emph{M}odel (DNPM).

Built upon DNPM, a novel Detailed3DMM is thus proposed, which augments conventional 3DMMs \cite{blanz1999morphable, facewarehouse, flame, multilinear3dmm} by introducing the synthesis of facial details only from the identity and expression inputs. Detailed3DMM is mainly based on the observation that, \emph{human facial wrinkles are mostly caused by various expressions of different individuals}. To capture such mapping, we elaborately designed an MLP-based encoder tasked with encoding the identity and expression vectors into the $w_+$ space \cite{abdal2019image2stylegan} of the pre-trained DNPM. Once the encoder is learned, it enables Detailed3DMM to be used like a traditional 3DMM which only takes identity and expression vectors as inputs, but also to synthesize detailed 3D face models \emph{without using image pixels}.

To demonstrate the effectiveness of the proposed DNPM and Detailed3DMM, we show that they can facilitate two downstream applications: speech-driven detailed 3D facial animation, and 3D face reconstruction from a degraded image. 1) Current speech-driven 3D facial animation methods \cite{FaceFormer2022, CodeTalker2023, Learn2Talk2024} can only generate a sequence of \emph{coarse 3D face models} from driving audio. The generation of a sequence of \emph{detailed 3D face models} from audio can yield more vivid and realistic animation, but it's really a challenging task as video is not involved. Thanks to Detailed3DMM, we can first construct an audio-expression regression network, and then feed the predicted expression vectors into Detailed3DMM to produce the detailed 3D facial animation. 2) In-the-wild facial images often have low resolution and quality due to unsatisfactory equipment or a low proportion of the whole scene. It's a nontrivial task to restore a detailed 3D face model from a limited amount of low-quality pixels. To achieve such a goal, we exploit the facial details prior provided by DNPM, by constructing CNN-based networks to regress the latent code in $w_+$ space from the degraded image. The regressed latent code is then inputted to the pre-trained DNPM to predict the high-resolution facial displacement map. We illustrate examples of the two applications in Fig. \ref{fig:banner}.

Finally, the contributions of this paper are summarized as follows.
1) The neural parametric model named DNPM for facial geometric details, which closes the gap between 3D face parametric representation and 3D face quality. 2) The Detailed3DMM, which has the advantage in the facial details synthesis from identity and expression, over traditional 3DMMs. 3) The new task, speech-driven detailed 3D facial animation, which can synthesize temporal facial details from driving audio but has not been explored by previous methods. 4) The novel parametric 3D face reconstruction method, which can restore a detailed 3D face model from a degraded image by mining the facial details prior from the pre-trained DNPM.

\section{Related Work}
\textbf{3D Face Parametric Model}. The first generic 3D face parametric model is built by Blanz and Vetter \cite{blanz1999morphable} using PCA on 3D face scans.
Since this seminal work, there has been a large amount of effort into improving the 3DMM modeling mechanism and constructing high-quality 3D face datasets. Paysan et al. \cite{3DBaselModel} propose the BaselFaceModel with 200 subjects in neutral expressions. To address the lack of expression modeling, Vlasic et al. \cite{multilinear3dmm} use a bilinear model to describe the combined effect of expression and identity variation on facial geometry. To provide more expression bases, FaceWarehouse \cite{facewarehouse} built a dataset captured from 150 subjects performing 20 expressions. To increase identity diversity, Booth et al. \cite{Booth1000} spent a large effort to build 3DMM from scans of 10K subjects.
Besides the modeling of shape and identity, Li et al. \cite{flame} add the parametrization of head and face poses and thus propose the FLAME model. The 3DMMs built on the above datasets can't well represent the facial details. FaceScape \cite{yang2020facescape} collects about 18K 3D face models with fine facial details, while FaceVerse \cite{faceverse} gathers about 2.7K detailed 3D face models of East Asians. Although the above two methods can predict facial details by the regression models trained on the datasets, they need additional RGB images as inputs. The main advantage of our proposed DNPM and Detailed3DMM is that they can synthesize facial details without using image pixels.

\textbf{Speech-Driven 3D Facial Animation.} Current SOTA methods, such as FaceFormer \cite{FaceFormer2022} CodeTalker \cite{CodeTalker2023} and Learn2Talk \cite{Learn2Talk2024}, usually train a regression network (e.g. Transformer) to learn the mapping from the speech audio features to the 3D facial motions. A limitation of the generated 3D facial motions is that they are in the form of coarse 3D face shapes, without enough facial details to represent fine-grained expressions and lip motions. In this paper, we propose a novel speech-driven detailed 3D facial animation method, which realizes the ability of synthesizing facial details from driving audio.

\textbf{Monocular 3D Face Reconstruction.} A common application of 3DMMs includes the estimation of the model parameters that best fit to a RGB image. One way is to conduct direct optimization in an analysis-by-synthesis framework \cite{blanz1999morphable, facewarehouse, flame, yang2020facescape}. Another way is to leverage the power of differentiable rendering to regress parameters from image data \cite{DECA, deng2019accurate, FaceRefiner}, without the need of 3D supervision. However, to the best of our knowledge, there is currently no parametric face reconstruction method that can recover a detailed 3D face model from a degraded image, e.g. blur and low resolution. Hence, our proposed 3D face reconstruction method based on DNPM fills this gap.

\section{Method}
The core of our DNPM is the StyleGAN v2 model \cite{karras2020analyzing}. We utilize about 7,000 face models' displacement maps in FaceScape dataset \cite{yang2020facescape} to train StyleGAN v2, and follow the suggested parameters in StyleGAN2-Ada \cite{karras2020training}. Once trained, 
we select $w_+$ space \cite{abdal2019image2stylegan} as the input parameters for DNPM. The $w_+$ space has many more degrees of freedom than the $w$ space \cite{karras2019style} and $z$ space \cite{karras2019style}, thereby being capable of reconstructing images well.

\textbf{$w_+$ space.} As the generator denoted by $\mathcal{G}$ in DNPM contains 18 layers of feature maps of resolution ranging from $4^2, 4^2, 8^2, 8^2, ...$ to $1024^2, 1024^2$, the latent code in $w_+$ space is defined by the concatenation of 18 different 512-dimensional $\mathbf{w}_i$ vectors. The generator $\mathcal{G}$ takes as input the latent code in $w_+$ space, and outputs a $1024^2$ displacement map:
\begin{equation}
     I = \mathcal{G}(\{ \mathbf{w}_i \}_{i=1}^{18}).
\end{equation}

\begin{figure}
\centering
\includegraphics[width=0.42\textwidth]{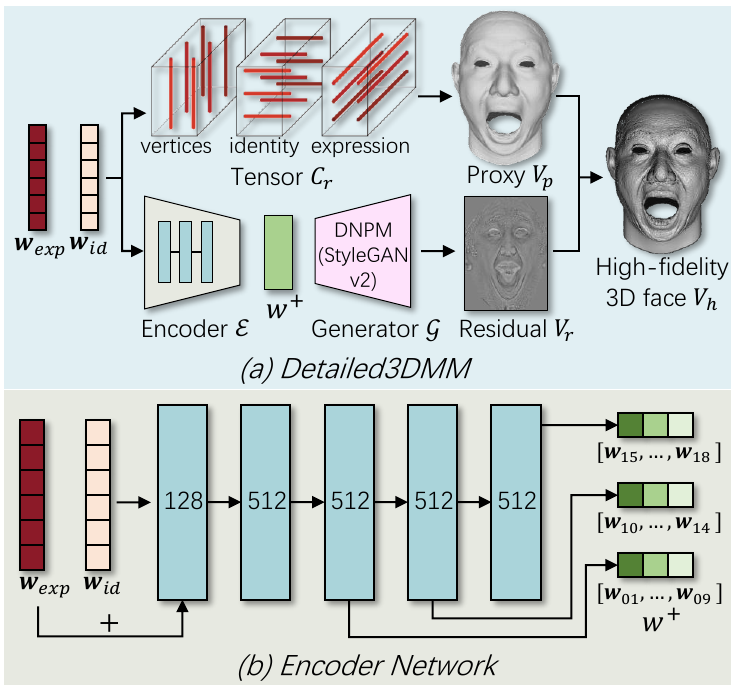}
\caption{Overview of the proposed DNPM and Detailed3DMM. The number in (b) encoder network denotes the number of channels in each layer.}

\label{fig:Detailed3DMM}
\end{figure}

\subsection{Detailed3DMM}
\label{sect:detailed3dmm}

\textbf{Formulation.} The ultimate goal of Detailed3DMM is to generate a detailed 3D face model $V_h$ only from the low-dimension identity vector $\mathbf{w}_{id}$ and expression vector $\mathbf{w}_{exp}$. As shown in Fig. \ref{fig:Detailed3DMM} (a), the generation of $V_h$ can further divide into the generation of proxy 3D face $V_p$ and the generation of residual 3D face $V_r$, which can be formulated as:
\begin{equation}
\label{vh}
    V_h = V_p + V_r.
\end{equation}
The proxy 3D face $V_p$ is generated by the widely used bilinear models \cite{multilinear3dmm}, given $\mathbf{w}_{id}$ and $\mathbf{w}_{exp}$:
\begin{equation}
\label{eq:vp}
    V_p = C_r \times \mathbf{w}_{id} \times \mathbf{w}_{exp},
\end{equation}
where $C_r$ denotes the core tensor related to the identity and expression basis. Meanwhile, the generation of the residual 3D face $V_r$ can be formulated by:
\begin{equation}
\label{eq:vr}
    V_r = \mathcal{T}'(s \cdot \mathcal{G}(\mathcal{E}(\mathbf{w}_{id}, \mathbf{w}_{exp}) + \overline{\mathbf{w}})),
\end{equation}
where $\mathcal{E}(\cdot)$ denotes the encoder mapping $\mathbf{w}_{id}$ and $\mathbf{w}_{exp}$ to the $w_+$ space of DNPM,
$\overline{\mathbf{w}}$ denotes the average vector in the $w_+$ space of DNPM, $\mathcal{G}(\cdot)$ denotes the generator in DNPM, $\mathcal{T}(\cdot)$ denotes the texture mapping function, from 3D coordinate space to 2D UV space, $\mathcal{T}'(\cdot)$ denotes the inverse texture mapping function, and $s$ is a scale parameter. As the generator $\mathcal{G}(\cdot)$ is pre-trained when constructing DPNM, the encoder $\mathcal{E}(\cdot)$ is then needed to be designed and trained to fulfill the building of Detailed3DMM.

\textbf{Encoder Design.} The encoder $\mathcal{E}(\cdot)$ is required to embed $\mathbf{w}_{id}$ and $\mathbf{w}_{exp}$ to the latent code in $w_+$ space of DNPM. We use MLP to construct the encoder, where all layers are standard fully-connected layers activated by ReLU. Motivated by network design in NeRF \cite{mildenhall2021nerf}, we add a skip connection that concatenates $\mathbf{w}_{exp}$ to the first layer's activation. By doing so, we can increase the differentiation between input data and thus help the network learn the mapping better. Following the spirit of pSp \cite{richardson2021encoding} that gets the output latent code from different feature maps, we obtain 9 style vectors from the 3rd layer, 5 style vectors from the 4th layer, and 4 style vectors from the 5th layer. The 18 style vectors are concatenated and fed into the generator $\mathcal{G}(\cdot)$. The network architecture of the encoder is shown in Fig. \ref{fig:Detailed3DMM} (b).

\textbf{Encoder Training.}
The encoder is trained with the paired data from the FaceScape dataset, in which each sample contains the $\mathbf{w}_{id}, \mathbf{w}_{exp}$ and their corresponding displacement map $I$.
The output of $\mathcal{G}$ is required to match the ground truth $I$. The training losses include the pixel reconstruction loss and the \emph{LPIPS} loss \cite{lpips}:
\begin{equation}
    \mathcal{L}_{pixel} = \parallel I - \mathcal{G}(\mathcal{E}(\mathbf{w}_{id}, \mathbf{w}_{exp}) + \overline{\mathbf{w}})\parallel_{1},
\end{equation}
\begin{equation}
    \mathcal{L}_{lpips} = \parallel \mathcal{F}(I)-\mathcal{F}(\mathcal{G}(\mathcal{E}(\mathbf{w}_{id}, \mathbf{w}_{exp}) + \overline{\mathbf{w}}))\parallel_{2},
\end{equation}
where $\mathcal{F}(\cdot)$ denotes the perceptual feature extractor based on  AlexNet \cite{krizhevsky2017imagenet}. The final loss for training the encoder in Detailed3DMM is defined as:
\begin{equation}
    \mathcal{L}_{d3d} = \lambda_{1} \mathcal{L}_{pixel} +\lambda_{2} \mathcal{L}_{lpips}.
\end{equation}
The hyper-parameters $\lambda_{1}, \lambda_{2}$ are set to 1, 0.05 respectively. 

\begin{figure*}
\centering
\includegraphics[width=0.95\textwidth]{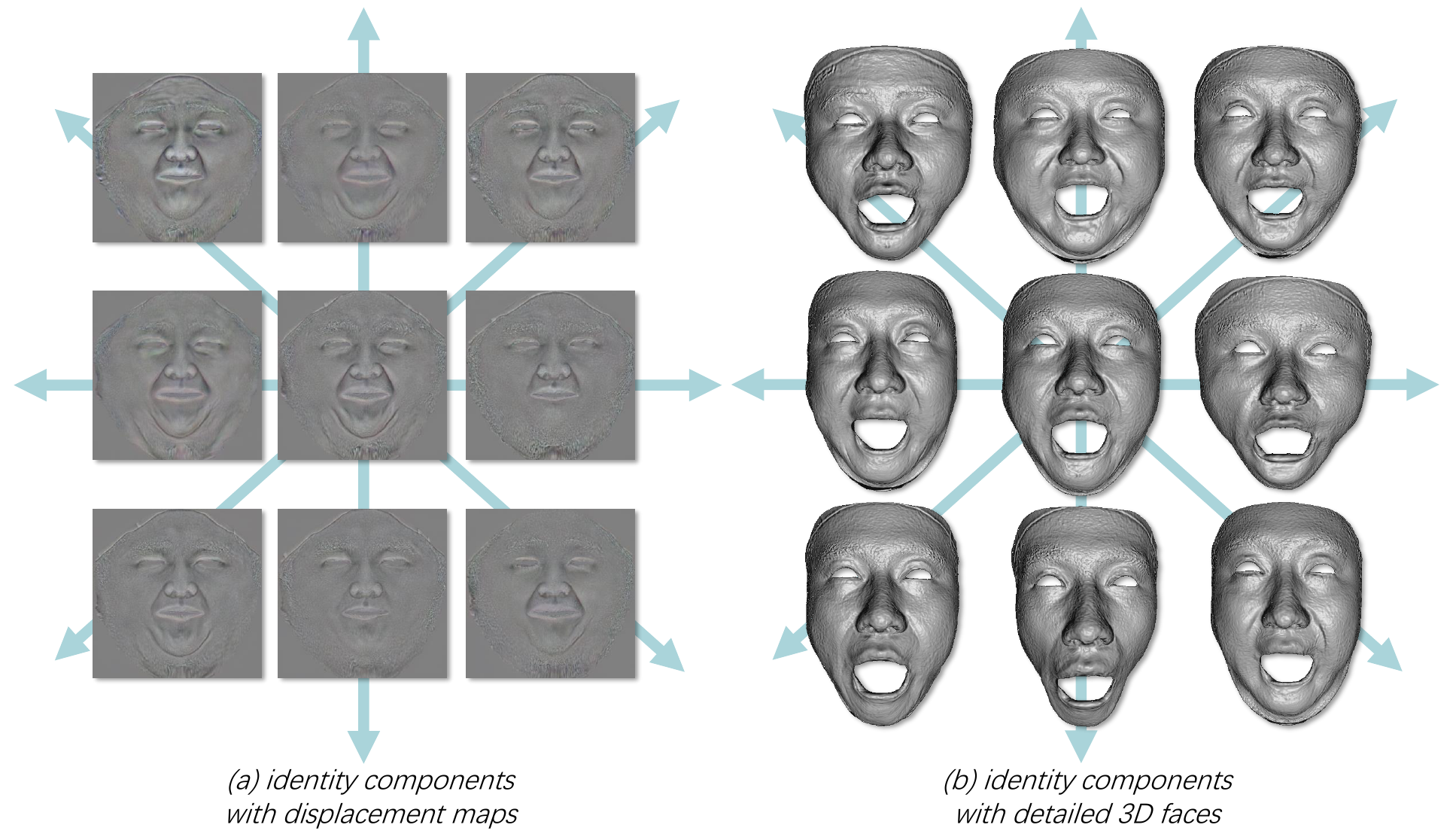}
\caption{Visualization of the identity principal components in the form of displacement map and detailed 3D face model.}
\label{fig:3DMMshow1}
\end{figure*}

\begin{figure*}
\centering
\includegraphics[width=0.98\textwidth]{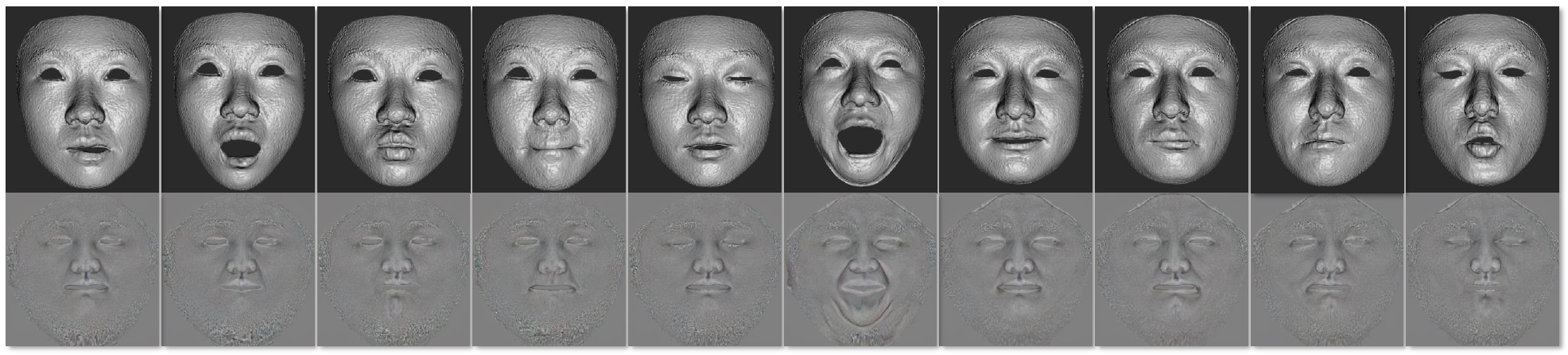}
\caption{Ten expression samples in the form of displacement map and detailed 3D face model.}
\label{fig:3DMMshow2}
\end{figure*}

\textbf{Visualization.} In Fig. \ref{fig:3DMMshow1}, we visualize the first four identity principal components augmented with the learned encoder, where the standard deviations are set to $\pm 200$. The faces in both displacement maps (Fig. \ref{fig:3DMMshow1} (a)) and 3D models (Fig. \ref{fig:3DMMshow1} (b)) maintain an almost consistent expression, despite the great identity changes. When modeling expression in Detailed3DMM, we don't perform PCA as the identity dimension. In Fig. \ref{fig:3DMMshow2}, we present ten expression samples with the same identity attribute. We can observe a similar phenomenon where the identity remains stable when expression varies.

\section{Applications}

\subsection{Speech-Driven Detailed 3D Facial Animation}
\label{sect:speech3danimation}

The goal of speech-driven detailed 3D facial animation is to synthesize both temporal proxy 3D faces $\{V_{p,t}\}_{t=1}^T$ and temporal residual 3D faces $\{V_{r,t}\}_{t=1}^T$ from the driving audio, the latter of which is added back to the former to produce the final detailed 3D animation $\{V_{h,t}\}_{t=1}^T$. 

We first train an audio-expression regression network. The network is based on the autoregressive model FaceFormer \cite{FaceFormer2022}, which contains two main modules, speech encoder and cross-modal decoder. The speech encoder adopts the pretrained speech model wav2vec 2.0 \cite{wav2vec2020}, while the cross-modal decoder uses self-attention to condition on past motions and uses cross-attention to condition on speech features. The expression vector is in the form of ARKit, which contains 52 elements. The training losses include the 3D vertex loss and the expression loss, which are defined as:

\begin{equation}
    \mathcal{L}_{vertex} = \sum_{t=1}^T||V_{p,t} - \hat{V}_{p,t}||_1,
\end{equation}

\begin{equation}
    \mathcal{L}_{exp} = \sum_{t=1}^T||\mathbf{w}_{exp,t} - \hat{\mathbf{w}}_{exp,t}||_1,
\end{equation}
where $\mathbf{w}_{exp,t}$ is the predicted expression vector in $t$-th frame, $V_{p,t}$ is obtained by inputting $\mathbf{w}_{exp,t}$ into the bilinear model defined in Eq. \ref{eq:vp}, $\hat{\mathbf{w}}_{exp,t}$ and $\hat{V}_{p,t}$ are the ground truth. The final loss for training the audio-expression regression network is defined as:
\begin{equation}
    \mathcal{L}_{speech} = \lambda_{1} \mathcal{L}_{vertex} +\lambda_{2} \mathcal{L}_{exp}.
\end{equation}
The hyper-parameters $\lambda_{1}, \lambda_{2}$ are set to 1, 0.01 respectively. 

Then, we use the trained regression network to regress the expression vectors $\{\mathbf{w}_{exp, t}\}_{t=1}^T$ from the inputted audio, and feed 
$\{\mathbf{w}_{exp, t}\}_{t=1}^T$ into the Detailed3DMM to generate 
$\{V_{p,t}\}_{t=1}^T$ by using Eq. \ref{eq:vp} and $\{V_{r,t}\}_{t=1}^T$ by using Eq. \ref{eq:vr}, finally obtaining $\{V_{h,t}\}_{t=1}^T$. As the identity vector can't be determined by the audio, it is selected by users and set the same for all frames.

\subsection{3D Face Reconstruction from a Degraded Image}
\label{sect:3DfaceDegraded}

Given a degraded RGB image, we first use a pre-trained bilinear face model to perform 3DMM fitting, in which the 3D landmarks on the bilinear face model are iteratively aligned with the estimated 2D landmarks in the image. Then, we use the pre-trained pix2pixHD network \cite{pix2pixHD} built on the FaceScape dataset \cite{yang2020facescape} to regress the displacement UV map from the inputted RGB image. Due to the low quality and resolution of the inputted image, the synthesized displacement UV map is often blurred and contains artifacts.
To realize the goal of restoring high quality displacement map $I_{high}$ from the low quality displacement map $I_{low}$, we construct a CNN-based encoder $\mathcal{E}_{deg}$ which maps $I_{low}$ to the $w_+$ space in DNPM:
\begin{equation}
    I_{gen} = \mathcal{G}(\mathcal{E}_{deg}(I_{low})+\overline{\mathbf{w}}).
\end{equation}

The training losses for supervising such a restoration network include the pixel reconstruction loss $\mathcal{L}_{pixel}$, the perceptual loss $\mathcal{L}_{lpips}$ and the identity consistency loss $\mathcal{L}_{id}$:
\begin{equation}
\label{eq:l1loss}
    \mathcal{L}_{pixel} = ||I_{gen} - I_{high} ||_1,
\end{equation}
\begin{equation}
\label{eq:lpiploss}
    \mathcal{L}_{lpips} = || \mathcal{F}(I_{gen}) - \mathcal{F}(I_{high})||_2,
\end{equation}
\begin{equation}
\label{eq:idloss}
    \mathcal{L}_{id} = \sum_{i=1}^5 (1-cos(R_{i}(I_{gen}),R_{i}(I_{high}))),
\end{equation}
where $\mathcal{F}(\cdot)$ denotes the perceptual feature extractor from VGG \cite{vgg}, $cos(\cdot, \cdot)$ measures the cosine similarity, and $R_i(\cdot)$ denotes the feature extracted from the $i$-th semantic level of the pre-trained face recognition network, ArcFace \cite{arcface}. The final loss for training the displacement map restoration network is defined as:
\begin{equation}
    \mathcal{L}_{deg} = \lambda_{1} \mathcal{L}_{pixel} + \lambda_{2} \mathcal{L}_{lpips} + \lambda_{3} \mathcal{L}_{id}.
\end{equation}
The hyper-parameters $\lambda_1, \lambda_2, \lambda_3$ are set to 10, 0.2, 0.1 respectively.

\section{Experiments}

\subsection{Implementation Details}

\textbf{DNPM}. About 7K displacement maps are selected from FaceScape \cite{yang2020facescape} and resized to $1024^2$ resolution to form the training dataset for DNPM. Two NVIDIA RTX 3090 GPUs are used to train DNPM for about 15K kimg, which is measured in thousands of real images shown to the discriminator. The total training time is about 15 days. 

\textbf{Detailed3DMM}. Another 7K displacement maps and their corresponding identity and expression vectors from FaceScape are used as the training dataset for Detailed3DMM. The MLP-based encoder $\mathcal{E}$ is trained for 100 epochs with Adam optimizer, in which the learning rate starts from 5e-4 in the first epoch and cosine decays in the rest epochs. The training batch size is set to 8. It takes about 14 hours to train $\mathcal{E}$ on a RTX 3090 GPU.

\textbf{Metrics.} In the quantitative evaluation, two metrics are used to measure the displacement map reconstruction quality: Peak Signal-to-Noise Ratio (PSNR) and Structural Similarity (SSIM). The higher the PSNR or SSIM, the better the reconstruction quality.

\begin{table}[]
\caption{Quantitative results of Detailed3DMM with different encoders over FaceScape dataset. The best score is colored by red, and the second-best score is colored by blue.}

\centering
\begin{tabular}{|l|l|l|c|}
\hline
Encoder & PNSR$\uparrow$    & SSIM$\uparrow$   & Num of Params$\downarrow$ \\ \hline
    $\mathcal{E}_{1700}$    & 24.050 & 0.588 & \textcolor{blue}{426.55M}       \\
    $\mathcal{E}_{5000}$    & \textcolor{red}{24.332} & \textcolor{red}{0.604} & 997.05M       \\
    $\mathcal{E}_{id,exp}$ (ours)    & \textcolor{blue}{24.083} & \textcolor{blue}{0.591} & \textcolor{red}{113.22M}       \\ \hline
\end{tabular}

\label{tab:ab_d3dmm}
\end{table}

\subsection{Ablation Study on Detailed3DMM}

We evaluate different methods of constructing the encoder in Detailed3DMM. We feed the identity and expression vectors into the 3DMM built on FaceScape to get a rough 3D face model, then downsample the 3D face model to 1.7K and 5K vertices. We train two encoders $\mathcal{E}_{1700}, \mathcal{E}_{5000}$ respectively to learn the mapping from the downsampled 3D face model to the latent code in $w_+$ space. These two encoders share a similar network with our encoder $\mathcal{E}_{id,exp}$ described in Sect. \ref{sect:detailed3dmm}, but have a different number of channels in fully connected layers. The quantitative evaluation results are reported in Tab. \ref{tab:ab_d3dmm}. Although the PSNR and SSIM scores of the three encoders are comparable, $\mathcal{E}_{id,exp}$ has the most lightweight network.

\begin{figure}
\centering
\includegraphics[width=0.47\textwidth]{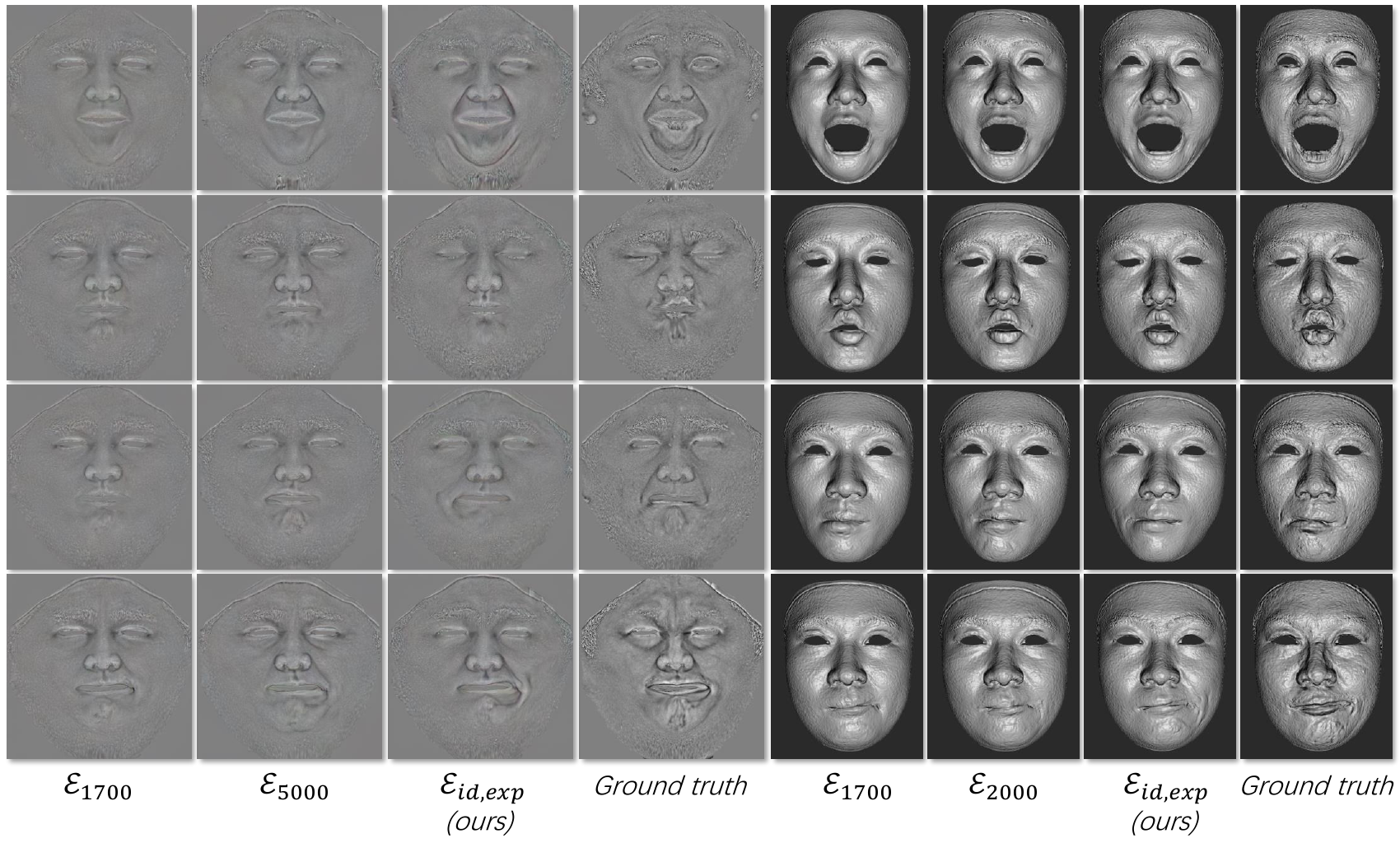}
\caption{Qualitative comparisons of different encoders over FaceScape.}
\label{fig:ab_d3dmm}
\end{figure}

 We qualitatively compare three different methods of constructing the encoder in Detailed3DMM, including $\mathcal{E}_{1700}, \mathcal{E}_{5000}$ and our encoder $\mathcal{E}_{id,exp}$. The qualitative comparisons are shown in Fig. \ref{fig:ab_d3dmm}. We observe that, firstly, with the increase in the number of vertices, more facial details related to expressions can be synthesized. Secondly, compared with both $\mathcal{E}_{1700}$ and $\mathcal{E}_{5000}$, the Detailed3DMM with $\mathcal{E}_{id,exp}$ can generate detailed 3D face models of more accurate expressions.
 
\subsection{Comparisons in Speech-Driven 3D Facial Animation}

We gathered about 4 hours of news report videos from the Internet to construct the audio-ARKit dataset, which is then used to train the Transformer-based regression network introduced in Sect. \ref{sect:speech3danimation}. The per-frame ARKit vector is predicted from the facial image by using a self-developed image-to-ARKit network. 

We conduct visual comparisons with two SOTA speech-driven 3D facial animation methods, FaceFormer \cite{FaceFormer2022} and CodeTalker \cite{CodeTalker2023}. The visual results are presented in the project page.  Note that we make comparison in the aspect of not lip-sync but detailed 3D animation. It can be observed that only our method can produce the animation of detailed 3D faces from the driving audio, while FaceFormer and CodeTalker can only produce rough 3D face sequences.

\subsection{Comparisons in Detailed 3D Face Reconstruction from a Degraded Image}
We train and test the displacement map restoration network on two datasets. 1) About 7K (training) + 0.2K (testing) displacement maps are selected from FaceScape, and then downsampled to form the paired high-low quality displacement maps dataset. 2) About 1K (training) + 0.1K (testing) frontal facial images of different ages and expressions are selected from FFHQ \cite{FFHQ}, and then downsampled to $64^2$ resolution and blurred by Gaussian smoothing, synthesizing the degraded images.

To make a quantitative comparison in restoring low-quality displacement map, we choose a optimization-based (PULSE \cite{menon2020pulse}) and two encoder-based (pSp \cite{richardson2021encoding}, E2Style \cite{wei2022e2style}) GAN inversion methods as the competitors. To make a fair comparison, all the competitors are re-trained on the above FaceScape and FFHQ datasets with the suggested parameters described in their papers. Additionally, we choose three SOTA 3D face reconstruction methods, DECA \cite{DECA}, FaceVerse \cite{faceverse} and FaceScape \cite{yang2020facescape}, to make extensive qualitative comparisons.

\begin{table}[]
\caption{Quantitative results in restoring 1/4, 1/8 downsampled displacement maps from FaceScape. s1 and s2 denote the 1st and 2nd stage in E2style.}

\centering
\begin{tabular}{|c|cc|cc|}
\hline
\multirow{2}{*}{Methods} & \multicolumn{2}{c|}{1/4}               & \multicolumn{2}{c|}{1/8}               \\ \cline{2-5} 
                         & \multicolumn{1}{c|}{PSNR$\uparrow$}    & SSIM$\uparrow$   & \multicolumn{1}{c|}{PSNR$\uparrow$}    & SSIM$\uparrow$   \\ \hline
PULSE                    & \multicolumn{1}{c|}{\textcolor{blue}{24.444}} & \textcolor{red}{0.670} & \multicolumn{1}{c|}{22.538} & 0.549 \\
pSp                      & \multicolumn{1}{c|}{23.124} & 0.562 & \multicolumn{1}{c|}{22.969} & 0.558 \\
E2style s1               & \multicolumn{1}{c|}{23.882} & 0.580 & \multicolumn{1}{c|}{\textcolor{blue}{23.627} } & \textcolor{blue}{0.570} \\
E2style s2               & \multicolumn{1}{c|}{23.816} & 0.576 & \multicolumn{1}{c|}{23.566} & 0.567 \\
Ours       & \multicolumn{1}{c|}{\textcolor{red}{24.524}} & \textcolor{blue}{0.611} & \multicolumn{1}{c|}{\textcolor{red}{24.551}} & \textcolor{red}{0.605} \\ \hline
\end{tabular}

\label{tab:4xSR}
\end{table}

\begin{table}[]
\caption{Quantitative results in restoring the degraded facial images from FFHQ.}

\centering
\begin{tabular}{|c|c|c|c|}
\hline
Methods      & PSNR$\uparrow$    & SSIM$\uparrow$   & Num of Params$\downarrow$ \\ \hline
PULSE        & 22.969 & 0.605 & -             \\
pSp          & 22.281 & 0.612 & 297.50M       \\
E2style s1   & 24.086 & 0.686 & \textcolor{blue}{141.61M}       \\
E2style s2   & \textcolor{blue}{24.086} & \textcolor{blue}{0.688} & 252.85M       \\
DNPM fitting & \textcolor{red}{25.687} & \textcolor{red}{0.707} & \textcolor{red}{114.27M}       \\ \hline
\end{tabular}

\label{tab:sr_ffhq}
\end{table}

\textbf{Comparions on Facescape.} 
We evaluate the restoring performance of different methods, on 1/4 and 1/8 downsampled displacement maps over FaceScape. As reported in Tab. \ref{tab:4xSR}, our method almost achieves the best PSNR and SSIM scores in both scaling factors. In the 1/4 scaling factor, our method outperforms all the encoder-based methods (pSp, E2Style) and achieves a considerable score compared with the optimization-based method (PULSE) which is quite time-consuming. As regards the 1/8 scaling factor, our method improves PSNR by 0.93 dB and SSIM by 3.5\% when compared with the second-best method E2Style.

\begin{figure}
\centering
\includegraphics[width=0.47\textwidth]{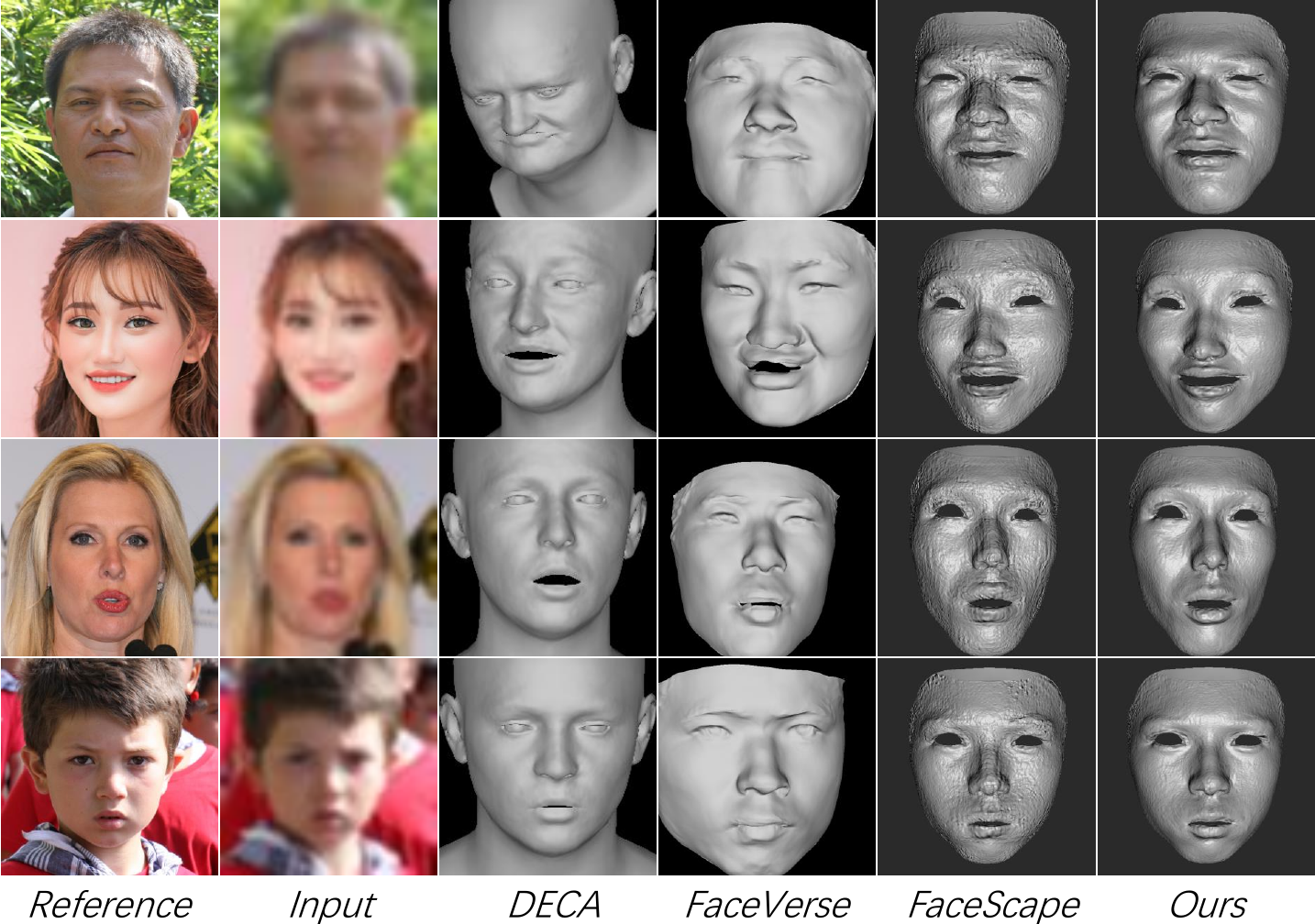}
\caption{The 3D face reconstruction results of different methods on the degraded facial images from FFHQ. }
\label{fig:sr_show_ffhq}
\end{figure}

\begin{figure}
\centering
\includegraphics[width=0.47\textwidth]{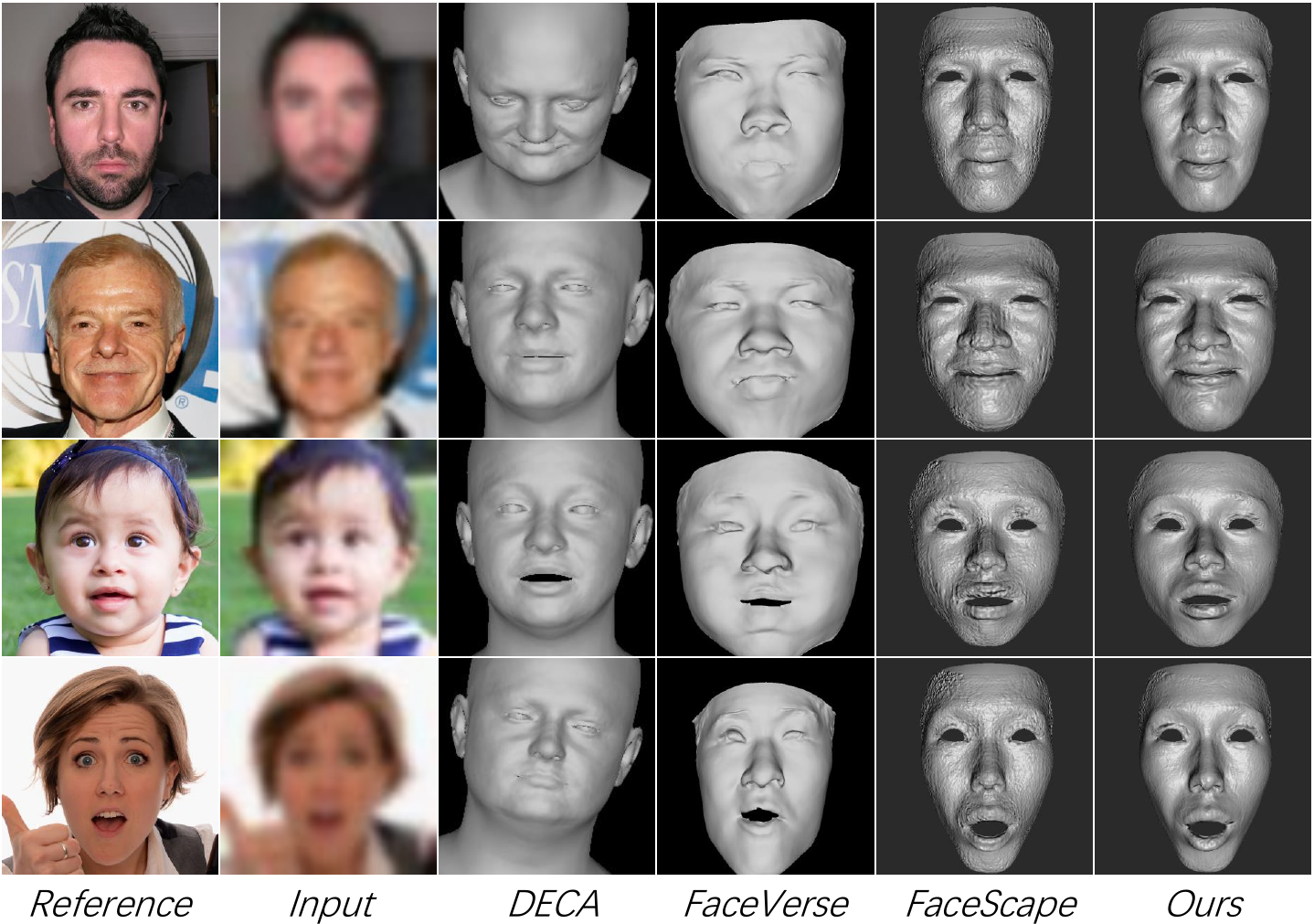}
\caption{The 3D face reconstruction results on degraded facial images from FFHQ.}
\label{fig:sr_show_ffhq1}
\end{figure}

\begin{figure}
\centering
\includegraphics[width=0.47\textwidth]{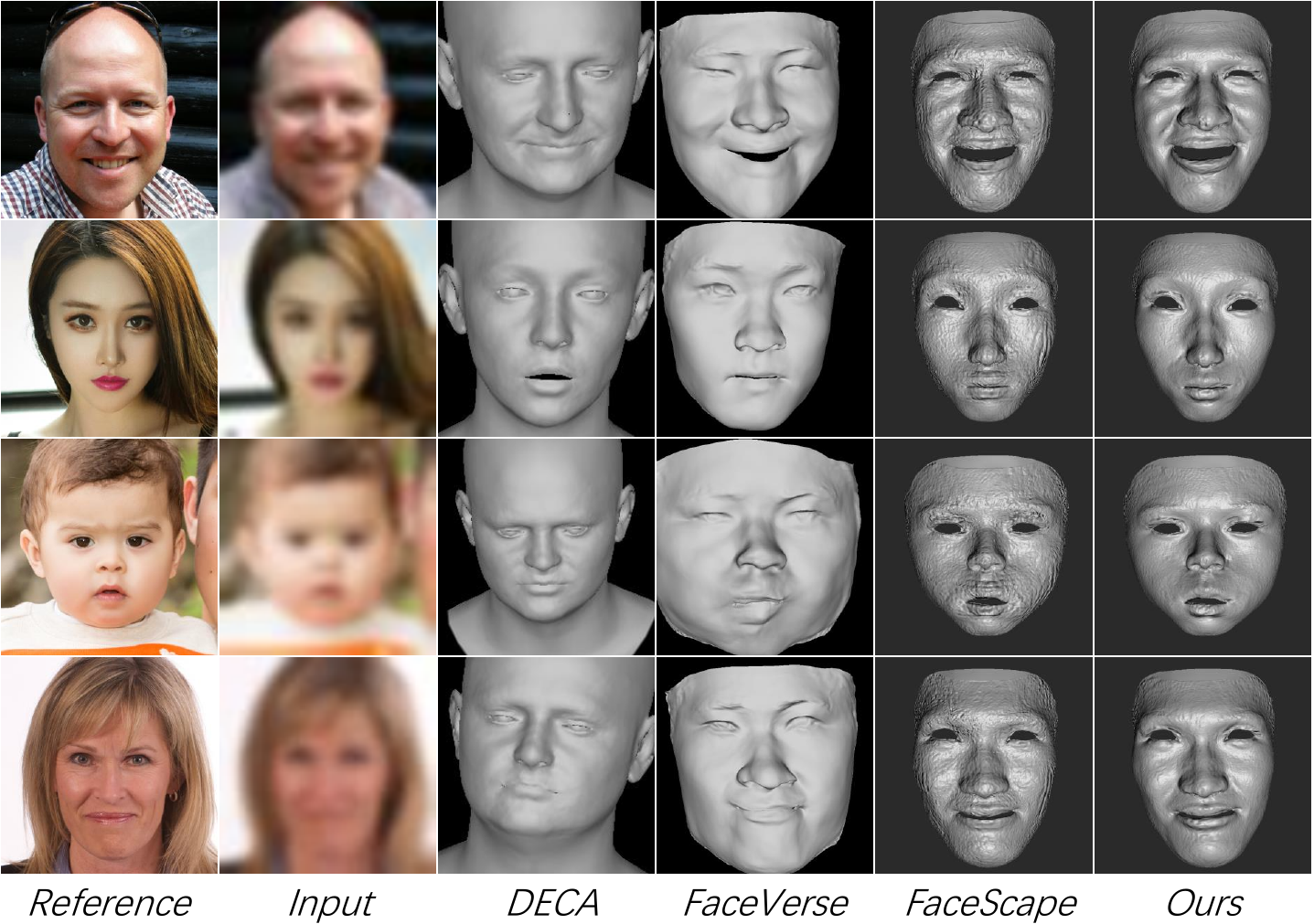}
\caption{The 3D face reconstruction results on degraded facial images from FFHQ.}
\label{fig:sr_show_ffhq2}
\end{figure}

\begin{figure}
\centering
\includegraphics[width=0.47\textwidth]{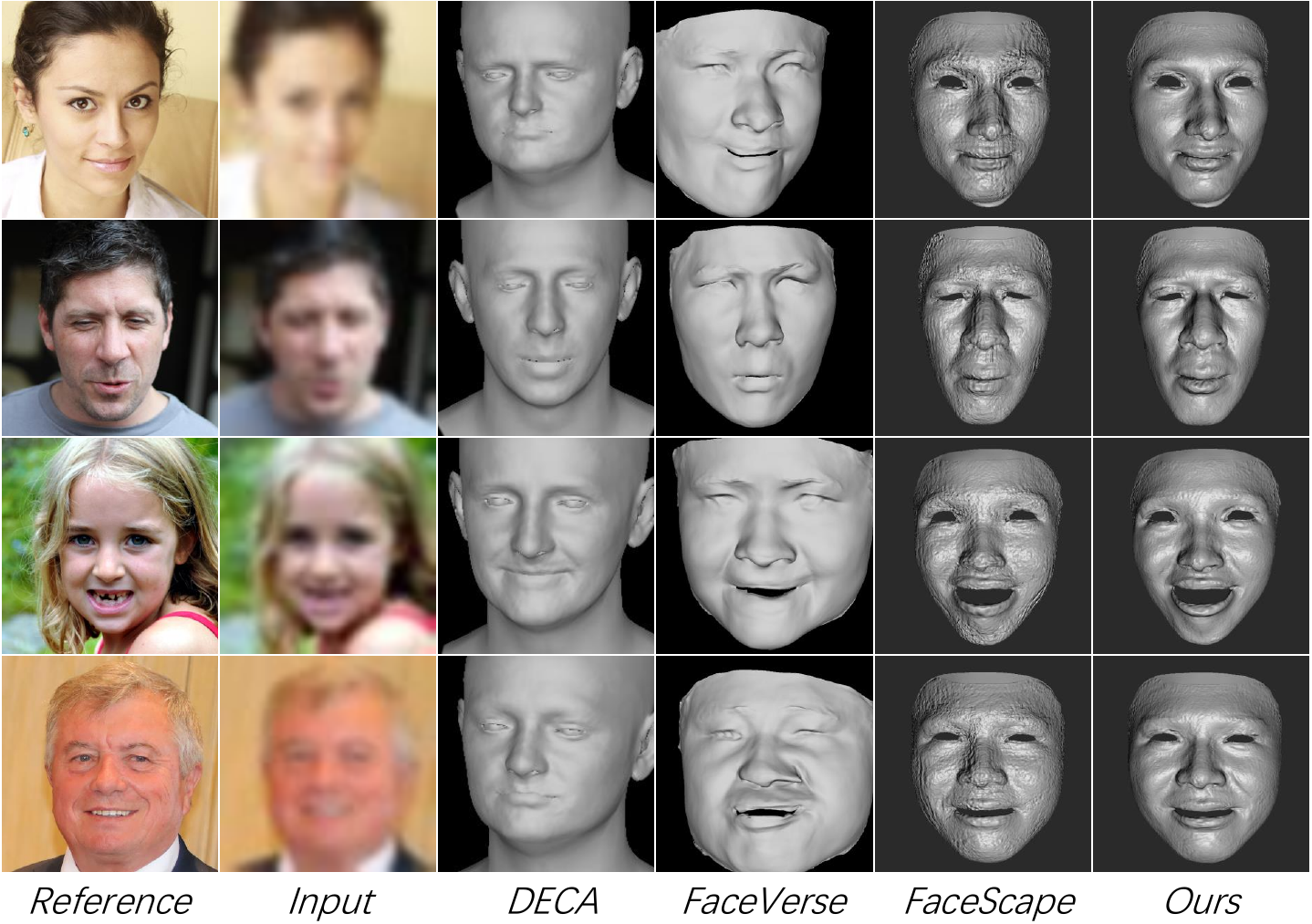}
\caption{The 3D face reconstruction results on degraded facial images from FFHQ.}
\label{fig:sr_show_ffhq3}
\end{figure}

\begin{figure}
\centering
\includegraphics[width=0.47\textwidth]{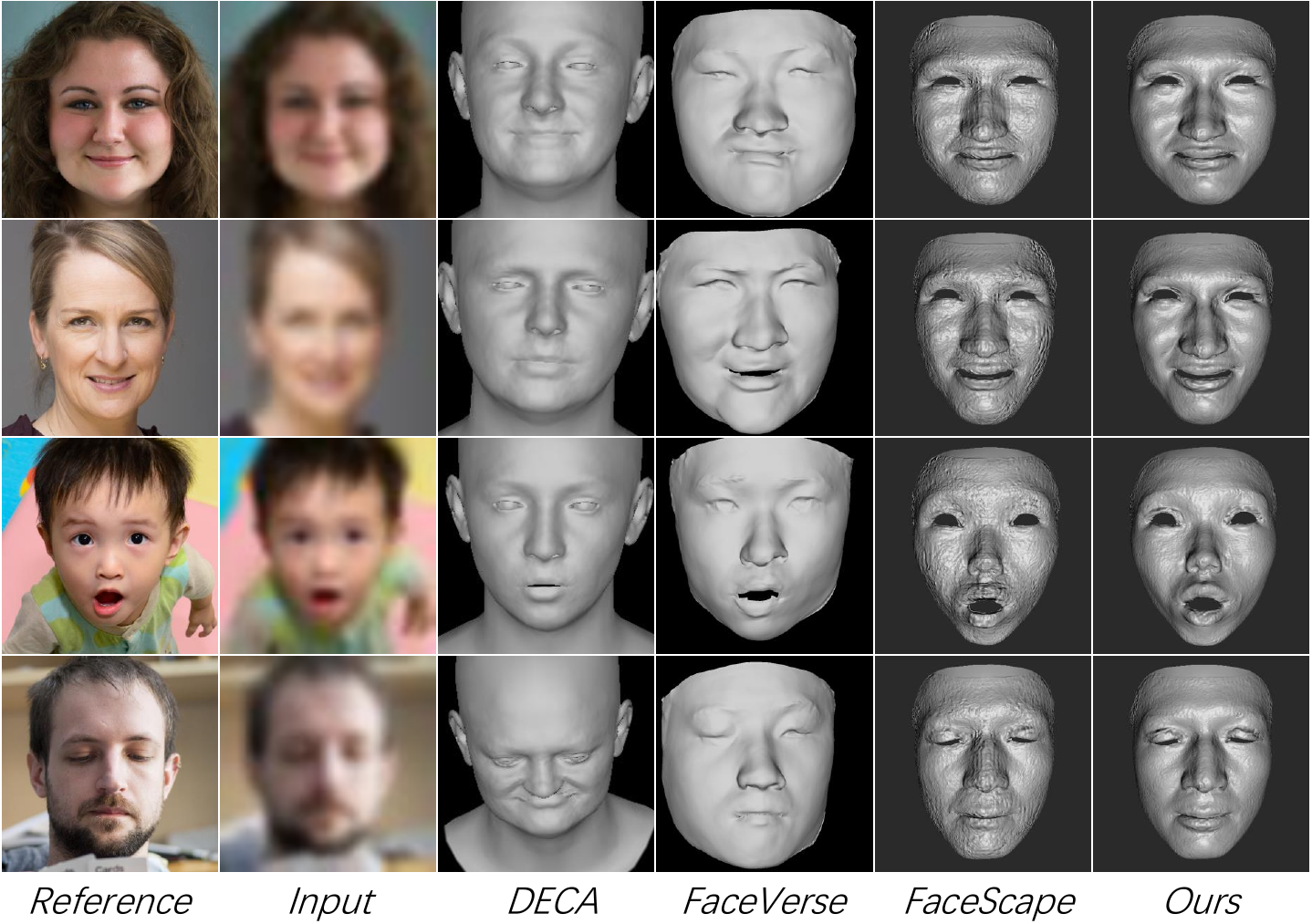}
\caption{The 3D face reconstruction results on degraded facial images from FFHQ.}
\label{fig:sr_show_ffhq4}
\end{figure}

\textbf{Comparisons on FFHQ.} 
We further evaluate the restoring performance of different methods on the degraded images from FFHQ. As reported in Tab. \ref{tab:sr_ffhq}, our method achieves the highest PSNR and SSIM scores, outperforming the second-best method E2style stage 2 by 1.60 dB in PSNR and 1.9\% in SSIM. Fig. \ref{fig:sr_show_ffhq},\ref{fig:sr_show_ffhq1}, \ref{fig:sr_show_ffhq2}, \ref{fig:sr_show_ffhq3} and \ref{fig:sr_show_ffhq4} shows the visual comparison among DECA, FaceVerse, FaceScape and our method. The reconstructed 3D faces by DECA and FaceVerse seem to have the proper expressions, but lack enough facial details, making the face identity differ from the reference images. The results of FaceScape suffer from unacceptable artifacts and unwanted wrinkles, such as stripes and grids, due to the blur in the inputted RGB image. Our method can generate satisfactory results in the aspects of rough shape, geometric details and face identity.

\begin{table}[]
\caption{Quantitative ablation results of removing different losses over FaceScape and FFHQ datasets.}

\centering
\begin{tabular}{|c|cc|cc|}
\hline
\multirow{2}{*}{Methods} & \multicolumn{2}{c|}{Facescape}     & \multicolumn{2}{c|}{FFHQ} \\ \cline{2-5} 
                         & \multicolumn{1}{c|}{PSNR$\uparrow$}   & SSIM$\uparrow$  & \multicolumn{1}{c|}{PSNR$\uparrow$}    & SSIM$\uparrow$   \\ \hline
w/o $\mathcal{L}_{id}$                   & \multicolumn{1}{c|}{24.181} & 0.607  & \multicolumn{1}{c|}{24.087}  & 0.694  \\
w/o $\mathcal{L}_{pixel}$                   & \multicolumn{1}{c|}{21.796} & 0.529 & \multicolumn{1}{c|}{22.338}  & 0.629  \\
w/o $\mathcal{L}_{lpips}$                   & \multicolumn{1}{c|}{\textcolor{red}{25.001}} & \textcolor{red}{0.655} & \multicolumn{1}{c|}{24.663}  & \textcolor{red}{0.738}  \\
Full Model             & \multicolumn{1}{c|}{\textcolor{blue}{24.524}} & \textcolor{blue}{0.611} & \multicolumn{1}{c|}{\textcolor{red}{25.687}}  & \textcolor{blue}{0.707} \\ \hline
\end{tabular}
\label{tab3}
\end{table}

\begin{figure}
\centering
\includegraphics[width=0.47\textwidth]{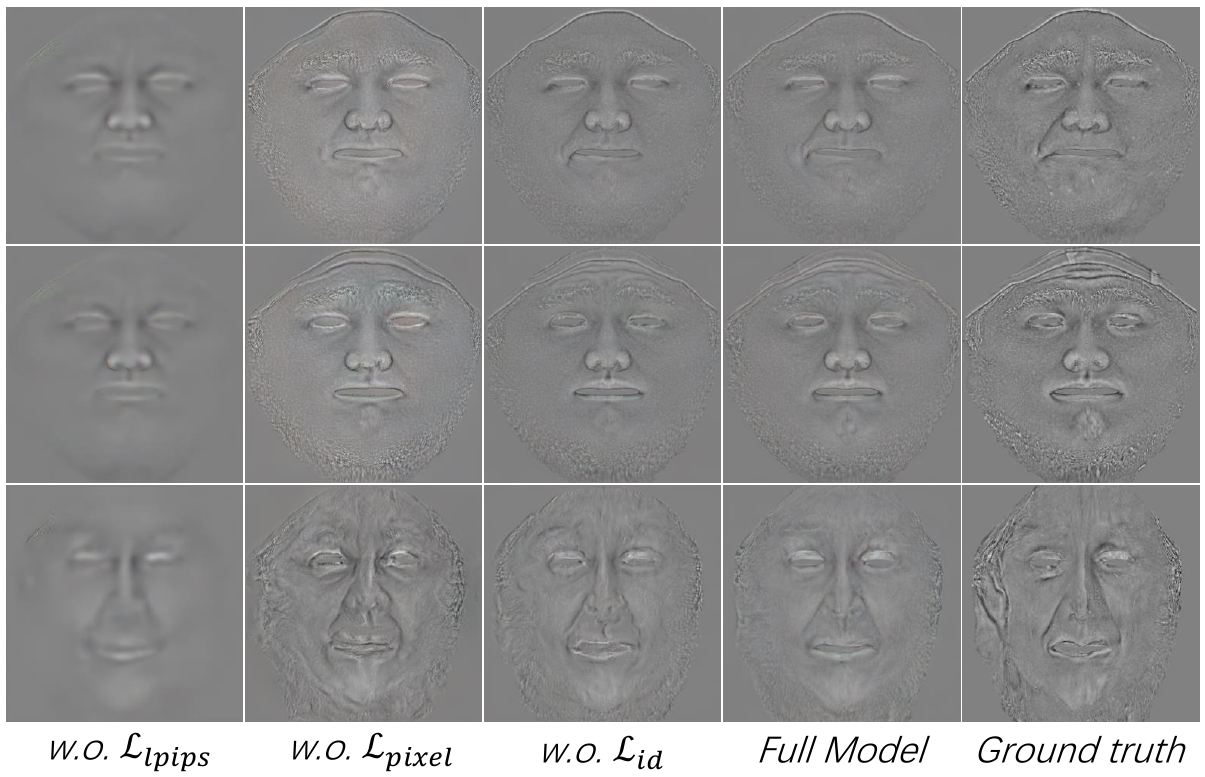}
\caption{The ablation results of removing different losses in training displacement restoration network. The 1st and 2nd rows show the restoration results on 1/8 downsampled displacement maps from FaceScape. The 3rd row shows the restoration results on the degraded facial image from FFHQ.}
\label{fig:em}
\end{figure}

\textbf{Ablation study.} We show the effectiveness of the three losses for supervising the restoration network: $\mathcal{L}_{pixel}$, $\mathcal{L}_{lpips}$ and  $\mathcal{L}_{id}$, by independently removing them from the overall loss.
Tab. \ref{tab3} reports the quantitative evaluation results on the testing datasets of both FaceScape and FFHQ. Removing $\mathcal{L}_{id}$ leads to a slight decrease in PSNR and SSIM, while removing $\mathcal{L}_{pixel}$ yields a large decrease in PSNR and SSIM. Although removing $\mathcal{L}_{lpips}$ can increase the performance in PSNR and SSIM, it generates very blurry displacement maps. The qualitative evaluation results are presented in Fig. \ref{fig:em}. It could be observed that, removing $\mathcal{L}_{id}$ introduces less identity consistency with ground truth, removing $\mathcal{L}_{pixel}$ yields abnormal pixel colors (1st, 2nd row) and artifacts (3rd row), and removing $\mathcal{L}_{lpips}$ results in very blurry displacement maps. 

\begin{figure}
\centering
\includegraphics[width=0.47\textwidth]{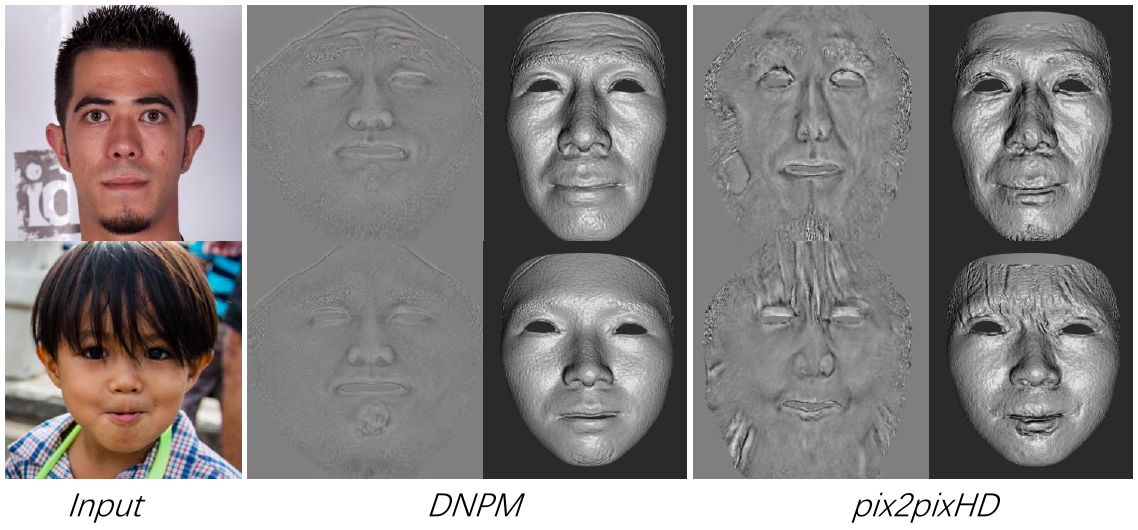}
\caption{Limitation: DNPM can't synthesize enough amount of facial details from high resolution RGB image, compared with pix2pixHD.}
\label{fig:limitation}
\end{figure}

\section{Conclusion and Limitations}
In this paper, we propose a neural parametric model named DNPM for facial geometric details modeling. Once learned, the high-fidelity facial geometric details can be represented by the low-dimension latent code. The proposed DNPM enables two downstream applications: speech-driven detailed 3D facial animation and 3D face reconstruction from a degraded image. As shown in Fig. \ref{fig:limitation}, a main limitation of DNPM is that it can't generate comparable amount of facial details with the image-to-displacement translation method, e.g. pix2pixHD \cite{pix2pixHD}, if high-resolution RGB image is inputted. DNPM is mainly used in situations where image pixels are not available or of small amount. To address this issue, our future work is to construct the parametric model on a much larger dataset. Another direction that is worth exploring is to replace the StyleGAN v2 with Diffusion Model \cite{StableDiffusion, DiffSFSR2024}.

\bibliographystyle{IEEEbib}
\bibliography{IEEE}

\end{document}